\documentclass[review]{elsarticle}
\graphicspath{ {./figures/} }
\usepackage{hyperref}
\usepackage{float}
\usepackage{verbatim} 
\usepackage{apalike}
\usepackage{color}
\usepackage{orcidlink}
\usepackage{amsmath}
\usepackage{diagbox}
\usepackage{multirow}                 
\usepackage{multicol}                 
\usepackage{multirow}                
\usepackage{float}                    
\usepackage{makecell}                 
\usepackage{amssymb}
\usepackage{tabularx}
\usepackage{booktabs}

\usepackage{algorithm,algpseudocode}
\floatstyle{plain}
\restylefloat{figure}

\journal{Computer Methods and Programs in Biomedicine}

\biboptions{authoryear}

\begin{document}
\begin{frontmatter}


\begin{titlepage}
\begin{center}
\vspace*{1cm}

\textbf{ \large IEFS-GMB: Gradient Memory Bank-Guided Feature Selection Based on Information Entropy for EEG Classification of Neurological Disorders}

\vspace{1.5cm}

Liang Zhang$^{a,b,*,\orcidlink{0009-0001-7272-5540}}$ (zl\_15556908539@126.com), 
Hanyang Dong$^{a,b,*,\orcidlink{0009-0000-7651-4065}}$ (dhy1042542469@126.com),
Jia-Hong Gao$^{c, d\orcidlink{0000-0002-9311-0297}}$(jgao@pku.edu.cn), Yi Sun $^{a,\orcidlink{0000-0003-1253-2381}}$ (sunyi@ucas.ac.cn),
Kuntao Xiao $^{a,\orcidlink{0009-0000-5121-7996}}$ (xiaokuntao@iai.ustc.edu.cn),
Wanli Yang $^{a,\orcidlink{0000-0002-0081-4376}}$ (yangwanli@iai.ustc.edu.cn),
Zhao Lv $^{e,**, \orcidlink{0000-0001-9727-366X}}$ (kjlz@ahu.edu.cn),
Shurong Sheng$^{a,**,\orcidlink{0009-0007-8116-2958}}$ (shengshurong@iai.ustc.edu.cn) \\

\hspace{10pt}

\begin{flushleft}
\small  
$^a$ Institute of Artificial Intelligence, Heifei Comprehensive National Science Center, Heifei, 230088, China. \\
$^b$ School of Artificial Intelligence, Anhui University, Heifei, 230601, China. \\
$^c$ McGovern Institute for Brain Research, Peking University, Beijing, China.\\
$^d$ Center for MRI Research, Academy for Advanced Interdisciplinary Studies, Peking University, Beijing, China.\\
$^e$ Anhui Province Key Laboratory of Multimodal Cognitive Computation, School of Computer Science and Technology, Anhui University, Heifei, 230601, China. \\ 
$^*$ These authors contribute equally to this work.\\
$^{**}$ Corrsponding authors.

\vspace{1cm}
\textbf{Corresponding author at: Institute of Artificial Intelligence, Heifei Comprehensive National Science Center, Heifei, 230088, China. } \\
Email: shengshurong@iai.ustc.edu.cn

\end{flushleft}        
\end{center}
\end{titlepage}

\title{IEFS-GMB: Gradient Memory Bank-Guided Feature Selection Based on Information Entropy for EEG Classification of Neurological Disorders}

\author[label1,label2]{Liang Zhang\corref{fir1}\orcidlink{0009-0001-7272-5540}} 
\ead{zl\_15556908539@126.com}

\author[label1,label2]{Hanyang Dong\corref{fir1}\orcidlink{0009-0000-7651-4065}}
\ead{dhy1042542469@126.com}

\author[label3,label4]{Jia-Hong Gao\orcidlink{0000-0002-9311-0297}}
\ead{jgao@pku.edu.cn}

\author[label1]{Yi Sun\orcidlink{0000-0003-1253-2381}}
\ead{sunyi@ucas.ac.cn}

\author[label1]{Kuntao Xiao\orcidlink{0009-0000-5121-7996}}
\ead{xiaokuntao@iai.ustc.edu.cn}

\author[label1]{Wanli Yang\orcidlink{0000-0002-0081-4376}}
\ead{yangwanli@iai.ustc.edu.cn}

\author[label4]{Zhao Lv\corref{cor1}\orcidlink{0000-0001-9727-366X}}
\ead{kjlz@ahu.edu.cn}

\author[label1]{Shurong Sheng\corref{cor1}\orcidlink{0009-0007-8116-2958}}
\ead{shengshurong@iai.ustc.edu.cn}

\cortext[fir1]{These authors contribute equally to this work.}
\cortext[cor1]{Corresponding authors.}
\address[label1]{Institute of Artificial Intelligence, Heifei Comprehensive National Science Center, Heifei, 230088, China.}
\address[label2]{School of Artificial Intelligence, Anhui University, Heifei, 230601, China.}
\address[label3]{McGovern Institute for Brain Research, Peking University, Beijing, China.}
\address[label4]{Center for MRI Research, Academy for Advanced Interdisciplinary Studies, Peking University, Beijing, China.}
\address[label5]{Anhui Province Key Laboratory of Multimodal Cognitive Computation, School of Computer Science and Technology, Anhui University, Heifei, 230601, China.}

\begin{abstract}
 
Deep learning-based EEG classification plays a pivotal role in the automated detection of neurological disorders, offering significant advantages in diagnostic accuracy and early intervention for personalized clinical treatment. However, the performance of such classification approaches is fundamentally limited by the intrinsic low signal-to-noise ratio characteristic of EEG signals. Consequently, feature selection (FS) is essential in optimizing the EEG representations derived from neural network encoders, thereby enhancing the overall efficacy of EEG classification frameworks.
Currently, few FS methods have been tailored for EEG neurological diagnosis, and most FS methods from other fields are designed for specific network architectures and lack clarity in interpretation, which restricts their direct utility in EEG classification. 
Moreover, existing FS methods predominantly rely on information from a single training iteration, which offers a narrow view of the data. Consequently, these approaches may lack the robustness necessary to effectively handle data variability. To address these challenges, we introduce IEFS-GMB, a novel \textbf{I}nformation \textbf{E}ntropy-based \textbf{F}eature \textbf{S}election approach guided by a \textbf{G}radient \textbf{M}emory \textbf{B}ank. This method begins by establishing a dynamic gradient memory bank that archives the sampled gradients from previous training iterations. Utilizing these gradients, it then calculates the information entropy to measure the importance of features. Finally, a feature weighting mechanism based on the calculated information entropy is applied to perform FS for EEG representations. Experimental outcomes across four public neurological disease datasets demonstrate that state-of-the-art EEG classification encoders enhanced with IEFS-GMB significantly outperform their counterparts without IEFS-GMB, with accuracy gains ranging from 0.64\% to 6.45\%. Superior performance is observed when comparing IEFS-GMB with four other FS techniques. Furthermore, our method increases the interpretability of the resulting model, an essential quality for its utility in medical applications. 
\end{abstract}


    \end{frontmatter}

\section{Introduction}
\label{introduction}

\begin{figure}
    \centering
    \includegraphics[width=0.5\linewidth]{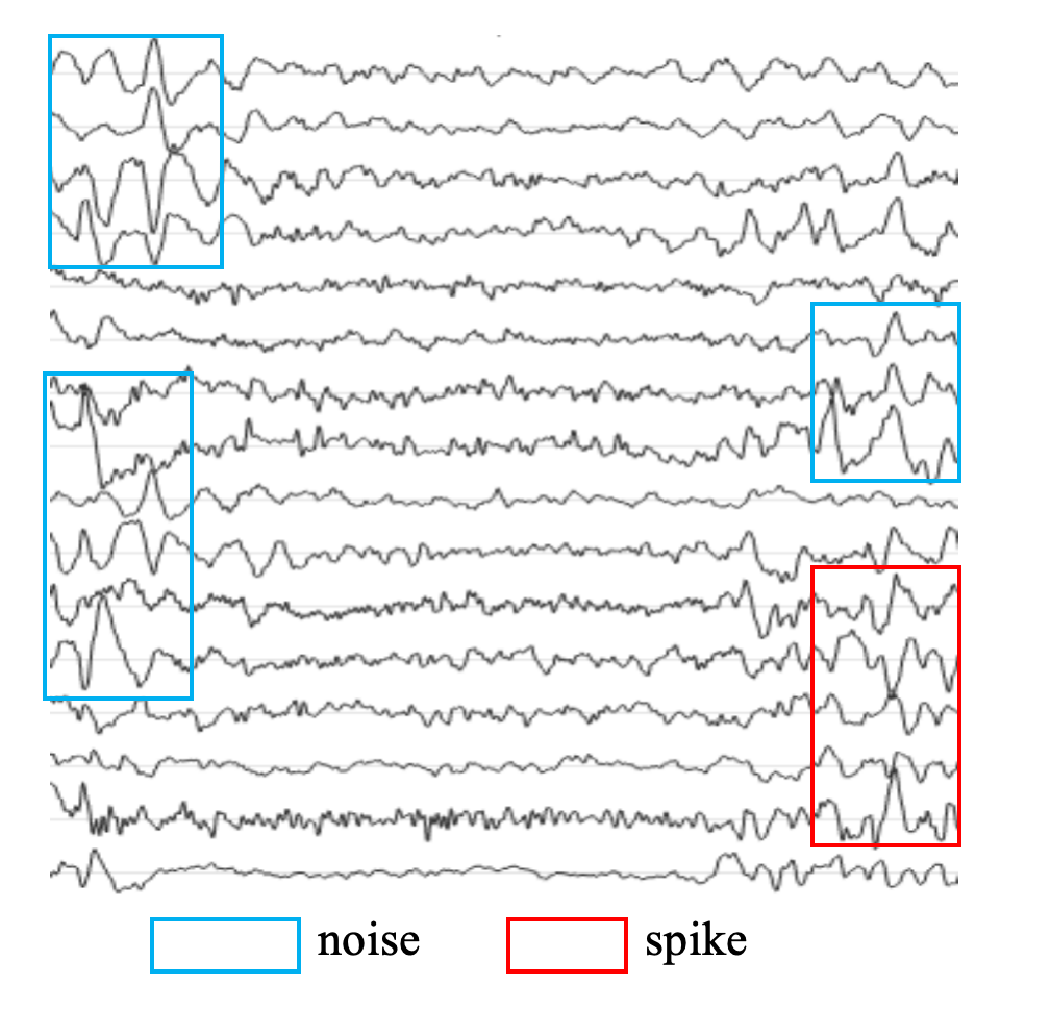}
    \caption{Preprocessed 16-channel EEG recording from the TUEV dataset, where noise artifacts mimic spike morphology, challenging automated epileptic spike detection.}
    \label{noisy_eeg}
\end{figure}

Electroencephalography (EEG) is a non-invasive and cost-efficient neuroimaging technique that records temporal fluctuations in the brain's electrical activity. Widely employed in clinical neurology, EEG plays a crucial role in diagnosing various neurological disorders, such as detecting epileptic seizures \cite{Ryan2024ElectrographicMF}, identifying interictal epileptiform discharges \cite{ref_satelight}, and screening for Parkinson's and Alzheimer's diseases \cite{JiaHCDZSS24}. Recent advances in deep learning have enabled the automation of EEG-based disorder classification, progressively replacing traditional manual interpretation methods \cite{Ranjan2024DeepLM}, thus improving diagnostic efficiency and objectivity. However, achieving high classification performance remains fundamentally constrained by the intrinsic low signal-to-noise ratio characteristic of EEG signals. As demonstrated in Figure \ref{noisy_eeg}, this persistent challenge exists even after rigorous preprocessing procedures \cite{Haghi2024EnhancedCO}, due to the nature of EEG as an electrical imaging modality \cite{meg_eeg_primer}. Therefore, it is paramount to develop robust feature selection (FS) techniques for various deep neural networks designed for EEG analysis.

However, the application of FS techniques specifically for deep learning-based EEG classification remains largely unexplored. Within this limited research landscape, existing FS methods suffer from narrow validation scopes, both in terms of model architectures and task domains. For instance, the TAL FS module incorporated in the CNN-based CLEP framework \cite{CLEP} has only been evaluated for seizure classification tasks. Similarly, the FS approach proposed by \cite{FS_eeg_RL} has been tested solely on EEGNet \cite{eegnet}, another CNN variant, leaving its generalizability to other architectures unverified.

While EEG-specific FS research remains scarce, the computer vision domain has demonstrated the remarkable potential of FS methods in enhancing image analysis performance. This success suggests promising transferability to EEG classification of neurological disorders. Notable examples include SCConv \cite{SCConv} and SAFM \cite{SAFM}, which have shown effectiveness on CNN architectures, and SMFA which has been validated on both CNNs and transformer-based ViT \cite{dosovitskiy2020vit}. However, these methods generally lack sufficient interpretability - a critical requirement for medical applications. Our work draws inspiration from the information entropy-based FS approach in \cite{Entropy_Pooling}, which offers several advantages particularly relevant to EEG analysis: (1) architectural flexibility beyond CNNs, (2) enhanced model interpretability through uncertainty measurement, and (3) dynamic feature weighting based on prediction confidence. The method evaluates feature significance by measuring the information entropy of network predictions associated with local feature vectors, where lower entropy indicates stronger feature-target correlation. However, its direct application to EEG faces two key limitations: (1) incompatibility with common EEG network architectures due to its reliance on global average pooling, and (2) restricted perspective from using only the previous iteration's activation map.

To overcome the challenges outlined above, we calculate the probability within the information entropy formula using the activation function of the heating map produced by Grad-CAM \cite{Grad_CAM}. Grad-CAM leverages the gradients of the target concept that flow into the final convolutional layer, generating a coarse heating map that underscores the pivotal regions in the image for predicting that concept. As a model visualization tool, Grad-CAM serves as a post-processing step. By integrating the probability computation approach introduced in Grad-CAM with the information entropy-based FS method proposed in \cite{Entropy_Pooling}, we create a plug-and-play FS module that can be seamlessly incorporated into various neural network architectures. Additionally, to broaden the data perspective accessible to the FS approach, we introduce a straightforward yet potent gradient memory bank. This memory bank enables us to dynamically preserve a record of sampled gradients from an extended history of iterations, beyond just the gradients from the immediate preceding iteration. The accumulated gradients are then employed in the calculation of probabilities within the information entropy formula. The key contributions of our paper are as follows:
\begin{itemize}
     \item We introduce a simple yet effective FS approach that integrates gradient-based probability computation, derived from Grad-CAM, with a prior information entropy-based FS method customized for CNNs. This combination results in a lightweight FS module versatile enough to be applied to various network architectures, while also offering high interpretability.
     
     \item We propose a novel gradient memory bank designed to dynamically preserve the historical gradients sampled from previous training iterations for the FS module. This memory bank expands the data perspective accessible to the FS module, thereby enhancing the robustness of the model to which our FS module is applied. 
     
     \item Extensive experiments have been conducted to validate the efficacy of the FS method we propose, named IEFS-GMB, across various neural network architectures and over four diverse datasets covering multiple tasks. These experiments also provide insights into the factors that affect its performance.
 \end{itemize}
The remainder of this paper is organized as follows. Section 2
reviews related research. Next, Section 3 describes our methodology. Section 4 illustrates the experiments, the evaluation metrics,
and discusses the results obtained by different models. Finally, Section 5 concludes this paper.

\section{Related Work}
\label{sec_related_work}

This section provides a comprehensive review of state-of-the-art neural network architectures for the classification of neurological disorders based on EEG, along with relevant multivariate time series classification models applicable to EEG analysis, hereafter referred to as ``EEG encoders". These architectures are particularly important as they serve as the foundational frameworks for integrating our proposed FS method. Subsequently, we systematically examine existing research focused on FS techniques.


\subsection{EEG Encoders}
EEG encoders are primarily built upon two architectures: CNNs and transformers. Among CNN-based approaches, EEGNet \cite{eegnet} and SpikeNet \cite{SpikeNet} are renowned as classic networks in EEG classification. Each network introduces a streamlined convolutional neural network that harnesses depthwise and spatial convolutions to encapsulate the distinctive features of EEG data. Different from the approach of using EEG time series as input in the aforementioned networks, \citeauthor{wu2023timesnet} performs EEG classification utilizing information of the frequency domain. It first decomposes a single time series into multiple non-overlapping 2D segments by principal frequency analysis. Subsequently, it utilizes inception-based modules to effectively capture the periodicity features within the EEG data. 

In the transformer category, Satelight \cite{ref_satelight} is an EEG classification model  that harnesses the power of both CNNs and transformers. The subsequent works discussed below are transformer-centric studies focused on multivariate time series analysis, which can be easily adapted for EEG classification. PatchTST \cite{PatchTST} begins by converting multivariate time series into independent univariate sequences.  Subsequently, it applies patch masking and reconstruction to these sequences using the standard vanilla Transformer architecture. This approach solely captures temporal dependencies, neglecting the correlations between different variables. To address this limitation, \citeauthor{ref_crossformer} introduced the Crossformer, which constructs a two-dimensional input representing time and variate dimensions. Crossformer then employs cross-time and cross-variate attention mechanisms to capture the intricate interdependencies within the input time series. In contrast to PatchTST and Crossformer, iTransformer \cite{ref_itransformer} is dedicated to modeling the relationships between variables, using the classic Transformer architecture to achieve state-of-the-art (SOTA) performance in various multi-time series prediction tasks. The ScaleFormer \cite{shabani2023scaleformer} and MTST \cite{MTST} models, meanwhile, investigate patching the input time series at multiple resolutions, diverging from the single-resolution approach of the previously mentioned studies. Medformer \cite{wang2024medformer}, conversely, integrates a multi-scale feature capturing strategy with cross-variate attention, enabling the model to learn both coarse-to-fine grained patterns in the temporal domain and the inter-variational correlations.

\subsection{Feature Selection Methods}
FS techniques for EEG data have traditionally been dominated by conventional machine learning methods, such as empirical mode decomposition \cite{Zheng2022ANF} and the Shapley value tree \cite{Yin2021AdaptiveFS}. These approaches may not be seamlessly integrable with deep learning models. The current landscape of FS methods tailored for deep learning-based EEG analysis is somewhat limited and has not been systematically assessed across various EEG encoders. For example, \citeauthor{FS_eeg_RL} utilized deep reinforcement learning (DRL) to distill pertinent information from EEG representations. However, their method is restricted to using EEGNet as the underlying architecture for embedding extraction from raw EEG data. Moreover, the efficacy of the DRL approach requires a considerable number of parameters. The TAL method \cite{CLEP} adopts a three-branch structure that leverages pooling and convolutional operations to select informative features across different dimensions of the CNN output.

FS strategies have demonstrated remarkable success in improving image analysis performance, presenting promising potential for adaptation to EEG encoder architectures For example, SCConv \cite{SCConv} incorporates two dedicated components: the Spatial Reconstruction Unit, which filters out unnecessary spatial details, and the Channel Reconstruction Unit, which sifts relevant data from among extraneous channel attributes.
The SMFA approach \cite{SMFA} employs a distinctive convolutional operation to enrich raw features, followed by a targeted extraction of both local and global information. By merging these complementary feature sets, the method successfully tackles the challenges in learning local features essential for super-resolution tasks. SAFM \cite{SAFM} employs parallel and autonomous convolutional processes that enable each head to handle varying scales of input information. These features are then aggregated to construct an attention map, facilitating spatially-adaptive feature modulation. This technique blends the efficiency of CNNs with the adaptability of transformers. While these methods are designed as plug-and-play, they lack the interpretability required in the EEG domain.
Information entropy-based FS, such as that proposed in \cite{Entropy_Pooling}, represents a transparent and modular technique; however, its applicability is confined to specific architectural frameworks, as outlined in the introduction.

\section{Method}
\label{sec_method}

\begin{figure}[ht]
    \centering
    \includegraphics[width=1\linewidth]{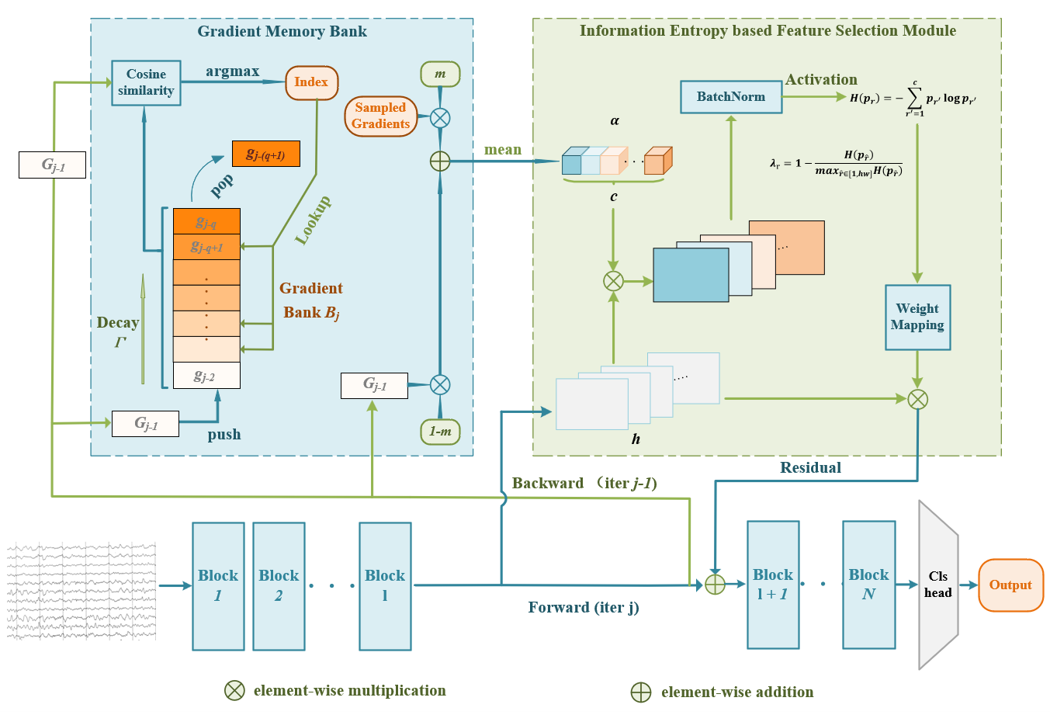}
    \caption{\textbf{Overview of the IEFS-GMB module integrated into an EEG encoder.} The whole workflow consists of three components: (1) Gradient memory bank which sample gradients from historical iterations; (2) Information entropy-based FS, leveraging the gradients captured in the gradient memory bank to create the feature weighting coefficient and perform FS; (3) Classification logits, employing features enhanced by the FS module for classification.}
    \label{fig_model_review}
\end{figure}

\subsection{Problem Definition}
Given an EEG encoder $\mathbb{E}$ , capable of generating the feature representation $\boldsymbol{h}$ for the input EEG clips $\{\boldsymbol{x}_i \in \mathbb{R}^{c \times t}\vert i=1,\cdots,N\}$ within a mini-batch, our objective is to apply the FS module IEFS-GMB to $\boldsymbol{h}$. This process refines the representation to $\hat{\boldsymbol{h}}$, making it cleaner and compact. The refined representation $\hat{\boldsymbol{h}}$ is then utilized to generate classification logits $\boldsymbol{y}$, thereby enhancing the overall classification performance of $\mathbb{E}$ on the entire dataset. Here $i$ signifies the sample index within a mini-batch. In this section, we assume all operations are performed in a specified layer $l$ of the encoder. The variables $c$ and $t$ associated with $\boldsymbol{x}_i$ represents the number of channels, and time duration respectively, of the EEG clip $\boldsymbol{x}_i$.

As depicted in Figure \ref{fig_model_review}, the workflow of our IEFS-GMB, integrated with the EEG encoder $\mathbb{E}$, comprises three primary components. The first component is the Gradient Memory Bank (GMB) module, which is responsible for sampling historical gradients from the EEG encoder. The second component is the information entropy-based FS module (IEFS), which leverages the bank produced by the GMB module to calculate the information entropy metric. This metric serves as an indicator of the relative importance of each local feature for the classification task and is subsequently employed for feature weighting. Finally, the refined feature representation derived from the IEFS module, is used to generate classification logits. We will provide a detailed explanation of these three components in the subsequent sections.

\subsection{Gradient Memory Bank}
\label{sec:gradientMB}

The GMB module introduced in this section draws inspiration from the previous study MoCo \cite{He_2020_moco}, which employs a queue mechanism to build a dynamic dictionary, thereby increasing the variety of negative samples available during mini-batch training. Additionally, it utilizes a momentum-based updating strategy for the key encoder to maintain its overall consistency as the training evolves within the framework of contrastive learning. To enable access to a richer history of gradients during mini-batch training, we have developed a comparable queue system. This system undergoes a three-step process to generate the final set of gradients, which are utilized for the subsequent calculation of information entropy.

\noindent\textbf{Step one: bank initialization}. Consider the current training iteration indexed as $j$, Let $\boldsymbol{h}_{i,j}$ represent the representation produced by layer $l$ of the EEG encoder $\mathbb{E}$ for each input clip $\boldsymbol{x}_i$. we initialize the gradient memory bank $\boldsymbol{B}_j$ with gradients from the preceding $q$ mini-batches up to the iteration before the last one as follows:
\begin{align}
    \boldsymbol{B}_j = \left[ \{\boldsymbol{g}_{i,j-(q+1)}\}, \cdots, \{\boldsymbol{g}_{i, j-2}\}\right] 
\end{align}
Here $\{\boldsymbol{g}_{i,j-(q+1)}\}$ represents the gradients obtained for each sample within mini-batch $j-(q+1)$ and $i$ ranges from 0 to $b$-1 with $b$ being the batch size. 

\begin{algorithm}[!t]
    \caption{Gradients sampling}
    \label{alg:tmg}
    \begin{algorithmic}[1]
    \Require Initialized Memory Bank $\boldsymbol{B}_j \in \mathbb{R}^{(q \times b) \times c \times h \times w}$.
    \Ensure Selected Gradients $\hat{\boldsymbol{B}}_j$ sampled from $\boldsymbol{B}_j$.
    \State Extract gradients from the last iteration as $\boldsymbol{G}_{j-1}$;  
    \For{gradients $\boldsymbol{g}_{i, j-1} \in \mathbb{R}^{c \times h \times w} $ within $\boldsymbol{G}_{j-1}$  }
        \For{gradients $\boldsymbol{g}_{i',g'} \in \mathbb{R}^{c \times h \times w}$  within $\boldsymbol{B}_j$ }
            \State $ sim(\boldsymbol{g}_{i', q'}, \boldsymbol{g}_{i,j-1}) =  \frac{\boldsymbol{g}_{i', q'}\cdot\boldsymbol{g}_{i, j-1}}{\Vert \boldsymbol{g}_{i', q'} \Vert_2 \cdot \Vert \boldsymbol{g}_{i, j-1} \Vert_2}$;
            \State $sim\_list.append(sim)$;
        \EndFor
        \State $top\_index = \mathbf{argTopK} (sim\_list) $
        \State $sampled\_gradients = \boldsymbol{B}_j[top\_index] $
        \State $\hat{\boldsymbol{B}}_j.push(sampled\_gradients)$
    \EndFor
    \end{algorithmic}
\end{algorithm}

\noindent\textbf{Step two: gradients sampling}. To curtail redundancy within the memory bank $\boldsymbol{B}_j$, we utilize a gradient sampling approach on its elements. Specifically, we pinpoint the top $K$ gradients that demonstrate the highest similarities to the gradients of the preceding iteration, with cosine similarity serving as as the criterion. This choice is grounded in the belief that the most recent gradients have the most significant effect. The methodology for selecting these gradients is outlined in Algorithm 1. In Algorithm 1, $\boldsymbol{G}_{j-1}$ is characterized by its dimensions of $(b,c,h,w)$, while $\boldsymbol{B}_j$ is of dimensions $(q\times b,c,h,w)$. The notation $\Vert\cdot\Vert_{2}$ refers to the $L2$-norm operation. The resultant set of selected gradients $\hat{\boldsymbol{B}}_{j}$ is dimensioned as $(b \times K, c, h, w)$. It is important to note that $K$ is a hyperparameter that governs the number of gradients retained for each sample in the previous iteration within iterations $j-(q+1)$ to $j-2$. The set of gradients $\hat{\boldsymbol{B}}_{j} \cup \boldsymbol{G}_{j-1}$ will be employed for subsequent computations. Furthermore, the retained gradients spans multiple past iterations after the gradient sampling process.

\begin{align}
    \boldsymbol{B'}_j &= \gamma \boldsymbol{G}_{j-1} \cup \gamma^{2}\{\boldsymbol{g}_{k_1, j-2}\}\cup \\ \notag
    & \cdots  \cup \gamma^{(q+1)}\{\boldsymbol{g}_{k_{q}, j-(q+1)}\}
\end{align}
\noindent\textbf{Step three: Assignment of a decay rate to the gradients}. Regarding the retained gradients $\hat{\boldsymbol{B}}_{j} \cup \boldsymbol{G}_{j-1}$ obtained in step two, we propose assigning a decaying rate to their constituent gradients based on the iteration index to which they belong. This is done under the assumption that the influence of gradients diminishes as the iterations from which they originate become more distant. The damping of gradient impact is governed by a hyperparameter $\gamma$, as outlined in Equation (2). 

It is worth noting that as the training progresses, the gradients of the current mini-batch are added to the queue, while the oldest gradients are dequeued to maintain the flow. This continuous turnover is the basis for our claim that the memory bank is inherently dynamic. 

\subsection{Information Entropy based Feature Selection}
\label{sec:gep}

In the FS module, the feature representation $\boldsymbol{h} \in \mathbb{R}^{b \times c \times h \times w}$ can be decomposed into a set of local features $\{\boldsymbol{h}_r \in \mathbb{R}^c\vert r=1,\cdots,hw\}$ after average pooling along the batch dimension. We start by computing the probability $\boldsymbol{p}_r$ for each $\boldsymbol{h}_r$ based on its corresponding heating map. Next, we introduce a weighting coefficient that inversely correlates with $\boldsymbol{h}_r$’s information entropy. This coefficient is then utilized to weight the heating map derived from the local feature $\boldsymbol{h}_r$, facilitating the FS process. The probability $\boldsymbol{p}_r$, which is crucial for entropy computation related to $\boldsymbol{h}_r$, can be derived through the following equations:
\begin{align}
    \boldsymbol{p}_{r} = Activation(\boldsymbol{v}_{r}) 
\end{align}
\begin{align}
   \boldsymbol{v}_r  = BN(\boldsymbol{\alpha} \cdot \boldsymbol{h}_r) 
\end{align}
\begin{align}
    \boldsymbol{\alpha} = m\cdot Avg(\boldsymbol{\tilde{B}}_{j}) + (1-m)\cdot Avg(\gamma \boldsymbol{G}_{j-1}) 
\end{align}
\begin{align}
    \boldsymbol{\tilde{B}}_{j} =\gamma^{2}\{\boldsymbol{g}_{k_1, j-2}\}\cup \cdots \cup \gamma^{(q+1)}\{\boldsymbol{g}_{k_{q}, j-(q+1)}\}
\end{align}

Here the $Activation(\cdot)$ can represent various activation functions such as $Sigmoid$ and $Softmax$ depending on the specific task at hand. The heating map $\boldsymbol{v}_r \in \mathbb{R}^c$ is computed as the batch normalization (BN) of the weighted feature map following Grad-CAM \cite{Grad_CAM}. The weights $\boldsymbol{\alpha}$ are determined by the gradients obtained from the previous section, adjusted by a momentum coefficient $m$. Both $Avg(\gamma \boldsymbol{G}_{j-1})$ and $Avg(\boldsymbol{\tilde{B}}_{j})$ have a dimension of $c$ since the $Avg$ operation denotes an average pooling to multiple dimensions of its input. 

Subsequently, we can create a feature weighting coefficient $\lambda_r$ for each heating map $\boldsymbol{v}_r$ with the following equations, in accordance with the information entropy-based FS approach proposed in \cite{Entropy_Pooling}. 
\begin{align}
    \label{Entropy}
    H(\boldsymbol{p}_{r})
    = - \sum_{r'=1}^{c} \boldsymbol{p}_{r'} \log \boldsymbol{p}_{r'}
\end{align}
\begin{align}
    \label{lambda_computation}
    \lambda_{r} 
    = 1 - \frac{H(\boldsymbol{p}_{r})}{\max_{\hat{r} \in [1, hw]}  H(\boldsymbol{p}_{\hat{r}})  }   
\end{align}

Ultimately, we perform FS to the set of local features $\{\boldsymbol{h}_r\}$ using the set of coefficients $\{\lambda_r\}$ with the linear weighting mechanism presented in Equation (9), and a residual connection is employed to produce the final features used for subsequent classification.
\begin{align}
    \hat{\boldsymbol{h}} = f_{IEFS} (\{\boldsymbol{h}_r\}) &= \{\lambda_{r} \cdot \boldsymbol{v}_{r}\} 
\end{align}
\begin{align}
      \boldsymbol{h}^{final} &= \{\boldsymbol{h}_r\} + \hat{\boldsymbol{h}}
\end{align}

\subsection{Classification Logits}
For the feature $\boldsymbol{h}^{final}$ enhanced by our FS framework, the subsequent classification process can be categorized into two cases: First, $\boldsymbol{h}^{final}$ is passed through the remaining part of encoder $\mathbb{E}_{res}$ if $\boldsymbol{h}^{final}$ refers to the representation from an intermediate layer of the encoder. Otherwise, a fully connected layer is directly applied on the flattened version of feature $\boldsymbol{h}^{final}$. Formally, the subsequent classification process can be described as follows:

\begin{align}
    \boldsymbol{h}^{out} &= \mathbb{E}_{res}(\boldsymbol{h}^{final}) 
\end{align}
\begin{align}
    \boldsymbol{y}  &= Linear(Flatten(\boldsymbol{h}^{out}))
\end{align}

\section{Experimental Results and Analysis}
\label{sec_results}
\subsection{Datasets and implementations}
\begin{table}[ht]
    \centering
    \scalebox{0.70}{
    \begin{tabular}{l|ccccccc}
        \Xhline{2pt}
        Datasets  & Subjects & Samples & Classes & Channels & Timestamps &Sampling Rates  \\
        \Xhline{1pt}
        TUEV     & 368     &13087  &2 &16 &250 &250Hz\\
        TUSZ     & 147    &27380   &2 &19  &400  &200Hz\\
        APAVA  & 23     & 5967  &2 &16  &256 &256Hz\\
        TDBRAIN & 72     & 6240 &2  &33  &256 &256Hz\\
        \Xhline{2pt}
    \end{tabular}
    }
    \caption{\textbf{Statistics of the four processed EEG datasets.} }
    \label{tab:datasets}
\end{table}

\noindent{\textbf{Datasets.}} To validate the effectiveness and robustness of the proposed IEFS-GMB, we conducted experiments on four publicly available datasets. Their descriptions and preprocessing procedures are as follows:
\begin{enumerate}
    \item [(1)]  \textbf{TUEV} \cite{TUEV_dataset} is the largest EEG dataset in the field of interictal epileptiform discharge (spike) detection, comprising samples annotated with binary labels indicating the presence or absence of epileptic spikes. We adhered to the processing procedures outlined in BIOT \cite{yang2023biot} and addressed several labeling errors in the dataset. Subsequently, we split the dataset into training, testing, and validation subsets, ensuring subject independence \cite{wang2024medformer}.
    
    \item [(2)] \textbf{TUSZ} \cite{TUSZ_dataset} consists of samples with binary labels for the diagnosis of epilepsy. We followed the procedures outlined in \cite{tang2022selfsupervised} and discarded samples with abnormal formatting. Subsequently, we split the dataset into training, testing, and validation subsets, ensuring subject independence.
        
    \item [(3)] \textbf{APAVA} \cite{APAVA_dataset} features samples marked with binary labels to denote the presence of Alzheimer’s disease.

    \item [(4)] \textbf{TDBRAIN} \cite{TDBRAIN_dataset} consists of samples that are tagged with labels signifying the diagnosis of Parkinson’s disease.

\end{enumerate}

The processing procedures and dataset splits for the final two datasets adhere to \cite{wang2024medformer}, employing a subject-independent validation strategy to assess model robustness across diverse patients. Table \ref{tab:datasets} summarizes the statistics of these datasets after preprocessing.

\noindent{\textbf{Baselines}} We integrated our IEFS-GMB module into the five leading EEG encoders, chosen from eight current SOTA options based on their results across the four benchmark datasets. These encoders include:
\begin{enumerate}
    \item [(1)] \textbf{TimeXer} \cite{TimeXer} enhances the traditional Transformer model by integrating the capacity to reconcile endogenous and exogenous variables. It employs patch-wise embedding for endogenous variables and variable-wise embedding for exogenous variables. With minimal alterations to the standard Transformer architecture, TimeXer demonstrates superior performance across various tasks. 

    \item [(2)] \textbf{iTransformer} \cite{ref_itransformer} challenges the conventional approach to temporal dimension embedding within the signal domain. It integrates the unified timestamp for each input channel into a solitary token and assesses the efficacy of this method across diverse transformer architectures. 

    \item [(3)] \textbf{MTST} \cite{MTST} applies the same embedding method as Crossformer. It designs a variable-length input module and a variable-length output module for irregular time-length inputs, enhancing the model’s predictive performance. 

    \item [(4)] \textbf{Crossformer} \cite{ref_crossformer} introduces a channel-independent signal embedding technique and employs a custom Router mechanism to facilitate an alternating self-attention mechanism across channel and temporal dimensions. 
    \item [(5)] \textbf{PatchTST} \cite{PatchTST} segments the original signal into non-overlapping patches along the temporal dimension for embedding. Each token is then processed through a vanilla transformer architecture, which yields improved prediction outcomes. 
    \item [(6)] \textbf{Satelight} \cite{ref_satelight} utilizes two depthwise convolutional layers, a pooling layer, and a batch normalization layer to derive embeddings from raw EEG data. Following this, the embeddings are fed into four self-attention modules. The prediction results are then obtained using a fully connected layer.
    \item [(7)] \textbf{SpikeNet} \cite{SpikeNet} is a convolutional neural network designed for modeling Interictal Epileptiform Discharges (IEDs). The network initially applies one-dimensional convolutions along the temporal axis, followed by one-dimensional convolutions along the spatial axis. Subsequently, the extracted features are processed through a series of residual modules, with each pair of blocks increasing the feature dimension by 32. 
    \item [(8)] \textbf{EEGNet} \cite{eegnet} is a convolutional neural network architecture that employs a convolutional kernel with a size corresponding to half the sampling rate, thereby emulating the conventional filter operation in electroencephalography. The network utilizes depthwise convolutional kernels to learn the frequency components following filtering, and it concludes with a separable convolution to integrate these features. 
\end{enumerate}
Additionally, we compared IEFS-GMB against four other SOTA FS methods:
\begin{enumerate}
    \item [(1)] \textbf{SAFM} \cite{SAFM} overcomes limitations in capturing long-range dependencies in high-resolution image reconstruction. It partitions features, applies depth-wise convolution, and integrates refined features. 
    \item [(2)] \textbf{SMFA} \cite{SMFA} addresses limitations in learning local information in super-resolution imaging. It enhances features with varying kernel sizes, extracts informative features, and fuses them. 
    \item [(3)] \textbf{SCConv} \cite{SCConv} filters convolutional features with \textbf{S}patial and \textbf{C}hannel \textbf{R}econstruction \textbf{U}nits. The SRU segments and recombines features to reduce spatial redundancy. The CRU dissects and fuses channel features using varying kernel sizes. 
    \item [(4)] \textbf{TAL} \cite{CLEP} is inspired by SENet and CBAM, the Triple Attention Layer (TAL) \cite{CLEP} enhances EEG features through a triple attention mechanism. This mechanism involves convolving the features, extracting and pooling salient aspects across three dimensions, concatenating these aspects, refining them, and finally summing them to enhance the feature representation. 
\end{enumerate}

\noindent{\textbf{Implementations and Evaluation Metrics.}} All experiments were performed using 6 RTX 4080 GPUs within a Python 3.9 virtual environment. The batch sizes were set to {64, 32, 32, 32} for the datasets TUEV, TUSZ, APAVA, and TDBRAIN, respectively, with the training epochs set to 200. During backward propagation, the Adam optimizer \cite{diederik2014adam} was employed, configured with a momentum of 0.9, a learning rate of 0.0001, and a weight decay of 0.0001. All experiments were initialized with the same random seed of 42 to ensure a fair comparison. In the Transformer-based approaches Crossformer, iTransformer, MTST, and TimeXer, the encoder comprised six layers, with a self-attention dimension of 128 for the attention-based layers, and a hidden dimension of 512 for the feed-forward networks. Eight heads are used for each attention-based layer within our attention mechanism. We utilized five evaluation metrics: accuracy, precision, recall, F1 scores, and AUROC.

\subsection{Experimental Results and Discussion}

\begin{table*}[htb]
    \centering
    \scalebox{0.48}{
    \begin{tabular}{c|c|ccccccccc}
        \toprule
        \multicolumn{2}{c|}{\diagbox{\textbf{Dataset}}{\textbf{Model}}} & \textbf{EEGNet} & \textbf{SpikeNet} & \textbf{Satelight} & \textbf{Crossformer} & \textbf{PatchTST} & \textbf{iTransformer} & \textbf{MTST} & \textbf{TimeXer} & \textbf{Ours} \\
        \midrule
        \multirow{4}{*}{\textbf{TUEV}} 
            & Acc & 83.91 & \textcolor{red}{90.78} & 80.47 & 86.08 & 90.42 & 84.40 & 88.25 & 82.91 & \textbf{93.58} \\
            & P/R & 70.20/\textcolor{red}{\textbf{96.24}} & \textcolor{red}{87.41}/92.23 & 66.37/92.98 & 74.65/92.98 & 82.34/93.48 & 79.49/93.78 & 77.73/94.49 & 70.51/90.48 & \textbf{94.07}/87.72 \\
            & F1 & 81.18 & \textcolor{red}{89.75} & 77.45 & 82.81 & 87.55 & 86.04 & 85.29 & 79.25 & \textbf{90.79} \\
            & AUC & 96.68 & \textcolor{red}{97.70} & 90.97 & 94.71 & 96.80 & 93.83 & 95.17 & 92.56 & \textbf{98.73} \\
        \midrule
        \multirow{4}{*}{\textbf{TUSZ}} 
            & Acc & 66.47 & 77.35 & 74.36 & 72.46 & 75.67 & 77.51 & 77.94 & \textcolor{red}{78.50} & \textbf{84.95} \\
            & P/R & 60.93/91.80 & 70.42/94.34 & \textcolor{red}{\textbf{88.95}}/55.64 & 68.36/83.64 & 77.81/71.83 & 71.86/90.43 & 70.72/\textcolor{red}{\textbf{95.35}} & 80.34/75.47 & 85.13/84.71 \\
            & F1 & 73.25 & 80.64 & 68.46 & 75.23 & 74.70 & 80.08 & \textcolor{red}{81.20} & 77.83 & \textbf{84.92} \\
            & AUC & 87.67 & \textcolor{red}{92.68} & 68.60 & 77.34 & 82.05 & 83.32 & 79.53 & 85.08 & \textbf{92.76} \\
        \midrule
        \multirow{4}{*}{\textbf{APAVA}} 
            & Acc & 78.69 & \textcolor{red}{79.40} & 74.50 & 71.88 & 69.74 & 74.79 & 71.24 & 67.97 & \textbf{80.04} \\
            & P/R & 75.24/95.57 & \textcolor{red}{75.66}/96.29 & 71.98/93.42 & 69.06/95.33 & 66.86/\textcolor{red}{\textbf{97.25}} & 71.84/94.62 & 69.05/93.42 & 65.69/96.41 & \textbf{77.29}/94.02 \\
            & F1 & 84.19 & \textcolor{red}{84.73} & 81.31 & 80.10 & 79.23 & 81.67 & 79.41 & 78.13 & \textbf{84.84} \\
            & AUC & 79.74 & 73.31 & 84.43 & 61.58 & 68.59 & \textcolor{red}{\textbf{87.28}} & 68.43 & 68.46 & 71.27 \\
        \midrule
        \multirow{4}{*}{\textbf{TDBRAIN}} 
            & Acc & 52.37 & \textcolor{red}{72.31} & 50.54 & 60.24 & 53.56 & 47.95 & 63.25 & 64.33 & \textbf{76.62} \\
            & P/R & 51.50/99.15 & \textcolor{red}{71.12}/76.12 & 50.54/\textcolor{red}{\textbf{99.98}} & 62.76/52.45 & 54.32/50.96 & 48.65/53.73 & 67.49/52.67 & 67.42/56.93 & \textbf{76.69}/77.19 \\
            & F1 & 67.78 & \textcolor{red}{73.53} & 67.14 & 57.14 & 52.58 & 51.06 & 59.16 & 61.73 & \textbf{76.94} \\
            & AUC & \textcolor{red}{\textbf{86.77}} & 80.86 & 55.66 & 63.41 & 61.76 & 47.63 & 67.26 & 70.94 & 84.49 \\
        \bottomrule
    \end{tabular}
    }
    \caption{\textbf{Comparison of Results with IEFS-GMB Integrated into Top-Performing EEG Encoders}. Bold figures indicate the highest performance achieved for each dataset, while numbers in red show the top performance of the eight EEG encoders prior to the integration of IEFS-GMB.}
    \label{tab_main_simple_results_1}
\end{table*}

\begin{table*}[tb]
    \centering
    \scalebox{0.65}{
    \begin{tabular}{c|c|ccccc}
        \toprule
        \multicolumn{2}{c|}{\diagbox{\textbf{Dataset}}{\textbf{Metric}}} & \textbf{Acc} & \textbf{P/R} & \textbf{F1} & \textbf{AUC}  & \textbf{Params (M)} \\
        \midrule
        \multirow{5}{*}{\shortstack{\textbf{TUSZ} \\ \textbf{(TimeXer)}}} 
     & \textbf{SCConv}   &81.61 &83.24/79.14 &81.14  &89.63  & 1196.08\\
     & \textbf{TAL}      &79.97 &\textbf{86.70}/70.80 &77.95 &91.00 & 1144.38 \\
     & \textbf{SMFA}     &77.86 &80.11/74.13 &77.00 &87.16 & 1273.56\\
     & \textbf{SAFM}     &77.66 &80.10/73.61 &76.72 &86.29  &1155.61\\
     & \textbf{IEFS-GMB (ours)}    &\textbf{84.95} &85.13/\textbf{84.71} &\textbf{84.92} &\textbf{92.76} &\textbf{1134.40}\\
    \midrule
    \multirow{5}{*}{\shortstack{\textbf{TUEV} \\ \textbf{(SpikeNet)}}} 
        & \textbf{SCConv}     &93.04   &85.78/\textbf{96.74}  & 90.93      &98.07     &573.83  \\
        & \textbf{TAL}     &\textbf{94.30}      &90.38/94.24    &\textbf{92.26}    &98.42     & 568.53\\
         & \textbf{SMFA}     &92.41    &87.41/92.23    &89.75  &97.71    &582.89 \\
         & \textbf{SAFM}     &93.31    &88.78/93.23       &90.95   &98.22   &568.31\\
         & \textbf{IEFS-GMB (ours)}     &93.58     &\textbf{94.07}/87.72   &90.79   &\textbf{98.73}   &\textbf{566.98}\\
    \midrule
    \multirow{5}{*}{\shortstack{\textbf{APAVA} \\ \textbf{(SpikeNet)}}} 
    & \textbf{SCConv}   &78.76  &75.21/95.81 &84.27  &\textbf{75.77}  &289.63 \\
    & \textbf{TAL}      &77.63 &75.72/91.75 &82.96  &73.79 & 287.17 \\
    & \textbf{SMFA}     &79.40 &75.66/96.29 &84.73  &73.31 & 286.86\\
    & \textbf{SAFM}     &75.28 &71.29/\textbf{97.73} &82.44  &68.25  &294.23\\
    & \textbf{IEFS-GMB (ours)} &\textbf{80.04} &\textbf{77.29}/94.02 &\textbf{84.84} &71.27 &\textbf{286.19}\\
        \bottomrule
    \end{tabular}
    }
    \caption{\textbf{Comparison with SOTA FS Methods.} Bold figures denote the peak performance reached for each dataset. ``Params. (M)", quantified in megabytes (MB), refers to the number of parameters when integrating FS methods into the deep layer of these backbones.}
    \label{tab_main_simple_results_2}
\end{table*}

\noindent{\textbf{Comparative Study When Integrating IEFS-GMB into SOTA EEG encoders.}} To measure the efficacy of our FS method IEFS-GMB, we start by testing eight SOTA EEG encoders across the four benchmark EEG datasets. Following this, we integrated IEFS-GMB with the top-performing encoders that demonstrated the highest accuracy for each dataset. For instance, IEFS-GMB was applied to SpikeNet on the TUEV dataset since SpikeNet performs best for this dataset. As illustrated in Table \ref{tab_main_simple_results_1}, the models enhanced with IEFS-GMB demonstrated improved performance in terms of both accuracy and F1 score when compared to the models without IEFS-GMB. The enhancements in accuracy ranged from 0.64\% to 6.45\%, and for the F1 score, the improvements spanned from 0.03\% to 3.72\%. The models augmented with our IEFS-GMB technique rank first in accuracy across the entire suite of four datasets. 
Furthermore, the robustness of IEFS-GMB in its integration with diverse architectures has been demonstrated. Notable performance enhancements are observed when IEFS-GMB is incorporated into the CNN-based SpikeNet model, as well as when it is integrated into the transformer-based models TimeXer and iTransformer.

\noindent{\textbf{Comparison to other FS methods.}} To assess the advancement of IEFS-GMB, we benchmark it against four leading SOTA FS techniques. The comparison is conducted using the top-performing EEG encoders: SpikeNet for the TUEV and APAVA datasets, as well as TimeXer for the TUSZ dataset. We assessed the performance of these methods by comparing their optimal outcomes at varying depths within the models. Specifically, for SpikeNet, which comprises 11 convolutional layers, we evaluated the 4th, 7th, and 11th layers to represent the shallow, middle, and deep layers, respectively. In the case of TimeXer, which has 6 attention layers, we conducted experiments on the 2nd, 4th, and 6th layers, corresponding to the shallow, middle, and deep layers.
As shown in Table \ref{tab_main_simple_results_2}, our method gets the best results in both accuracy and F1 score on most datasets with the fewest parameters. On the TUSZ dataset, IEFS-GMB surpasses the top-performing baseline, SCConv, by a margin of 3.34\% in accuracy. Similarly, on the AVAPA dataset, IEFS-GMB edges out the leading baseline, SMFA, with a 0.64\% accuracy improvement. For the TUEV dataset, IEFS-GMB ranks as the second most effective method.

\begin{table*}[!tb]
    \centering
    \scalebox{0.7}{
    \begin{tabular}{c|c|cccc}
        \toprule
        \multicolumn{2}{c|}{\diagbox{\textbf{Model}}{\textbf{Metric}}} & \textbf{Acc} & \textbf{P/R} & \textbf{F1} & \textbf{AUC} \\
        \midrule
        \multirow{4}{*}{\shortstack{\textbf{TUEV} \\ \textbf{(SpikeNet)}}} 
            & \textbf{w/o IEFS-GMB}  & 90.78  & 87.41/92.23 & 89.75 & 97.70 \\
            & \textbf{Deep Layer}    & 92.86  & 91.45/88.47 & 89.94 & 97.84 \\
            & \textbf{Middle Layer}  & \textbf{93.58} & \textbf{94.07} & 87.72/\textbf{90.79} & \textbf{98.73} \\
            & \textbf{Shallow Layer} & 92.77  & 88.07/\textbf{92.48} & 90.22 & 98.14 \\
        \midrule
        \multirow{4}{*}{\shortstack{\textbf{TUSZ} \\ \textbf{(TimeXer)}}} 
            & \textbf{w/o IEFS-GMB} & 78.50  & 80.34/75.47 & 77.83 & 85.08 \\
            & \textbf{Deep Layer}   & \textbf{84.95} & 85.13/\textbf{84.71} & \textbf{84.92} & \textbf{92.76} \\
            & \textbf{Middle Layer} & 81.16  & \textbf{85.53}/75.02 & 79.93 & 91.57 \\
            & \textbf{Shallow Layer}& 83.49  & 84.33/82.26 & 83.28 & 92.05 \\
        \bottomrule
    \end{tabular}
    }
    \caption{\textbf{Experimental results when integrating our IEFS-GMB to different layers within SpikeNet on TUEV and TimeXer on TUSZ. Bold numbers denote the best results.}}
    \label{tab_robust_simple_results_2}
\end{table*}

\noindent{\textbf{Robustness Studies on Different Layers.}}
Table \ref{tab_robust_simple_results_2} showcases the outcomes of integrating IEFS-GMB into the shallow, intermediate, and deep layers of the top-performing models SpikeNet and TimeXer, across the TUEV and TUSZ datasets. Clear improvements are observed in this table, validating the robustness of our method as a flexible plug-and-play module, capable of integration with various layers of an EEG encoder. 

\begin{figure*}[tb!]
\centering
\includegraphics[width=1.0\linewidth]{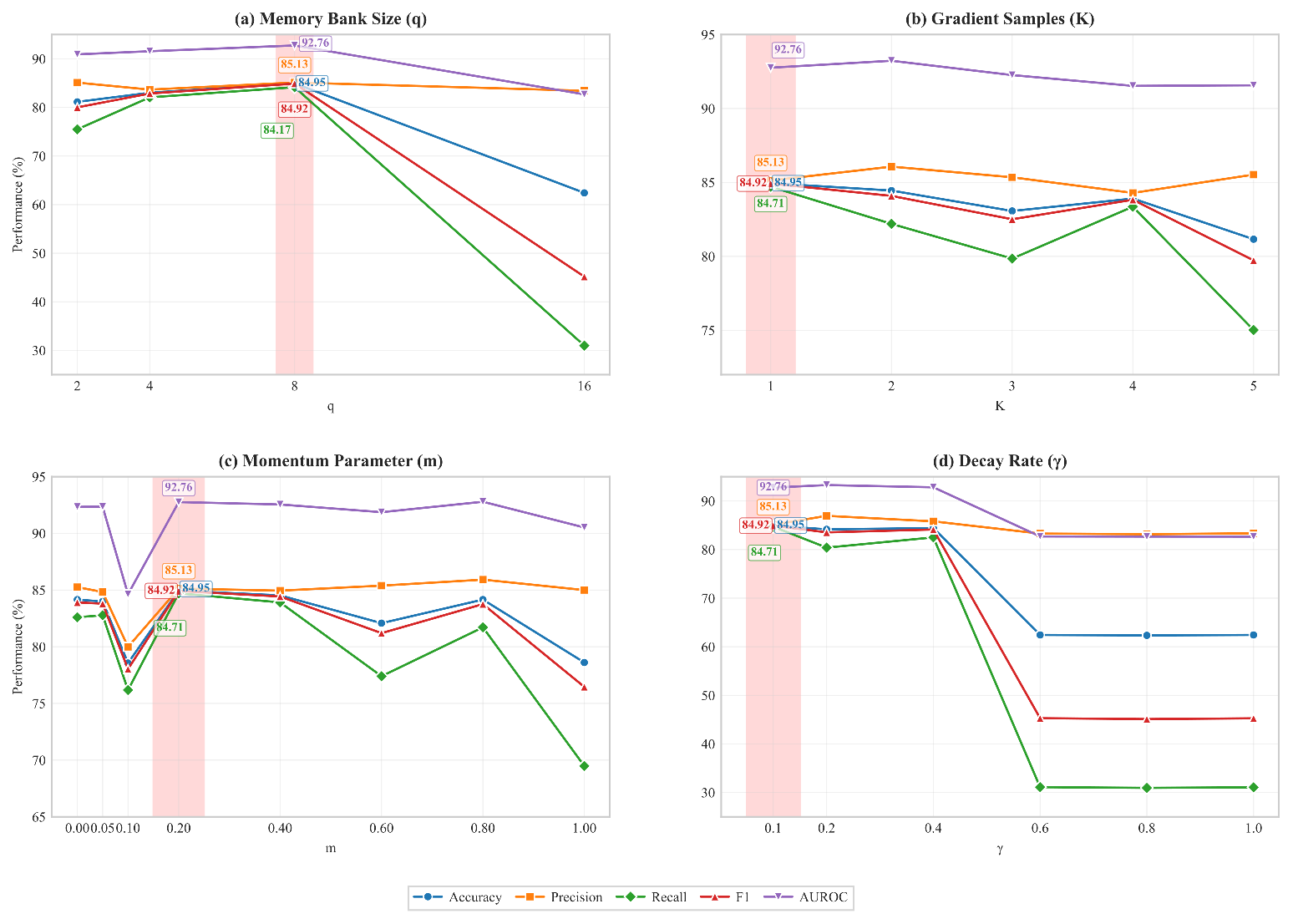}
    \caption{\textbf{Hyperparameter experiments of IEFS-GMB in TimeXer framework for EEG classification on the TUSZ Dataset.} Four hyperparameters of the IEFS-GMB approach are examined: memory bank size $q$, the number of gradient samples $K$, the momentum parameter $m$ for the gradient memory bank, and the decay rate $\gamma$. The red-shaded area indicates the optimal region where accuracy serves as the primary performance metric.
    }
    \label{fig_ablation_studies}
\end{figure*}

\noindent{\textbf{Ablation Study on the Hyperparameters. }}
To examine the impact of hyperparameters within the IEFS-GMB module, we conducted ablation studies using the top-performing EEG encoder TimeXer on the TUSZ dataset. We selected TUSZ for the ablation study because of its substantial data volume.

As shown in Figure \ref{fig_ablation_studies}, 8 is the optimal value regarding the size of gradient memory bank $\mathbf{q}$. Concerning the hyperparameter $K$, which signifies the number of gradients retained for each sample from the previous iteration, the optimal value is 1. The model is notably sensitive to the momentum parameter $m$. Performance notably declines when $m$ is either excessively low or too high. When $m$ is assigned a value of 0, the IEFS module relies solely on gradients from the most recent iteration for FS. Conversely, when $m$ is set to 1, the module incorporates gradients from earlier iterations, excluding those from the latest iteration. In both scenarios, the model’s performance is inferior to when $m$ is set to 0.2, thereby confirming the efficacy of utilizing a gradient memory bank to sample and retain gradients from multiple past iterations. Regarding the decay rate $\gamma$, the model’s performance remains relatively consistent when $\gamma$ is below 0.5. However, when $\gamma$ exceeds 0.5, there is a notable performance decline. Since a larger $\gamma$ implies a greater influence of gradients from more remote iterations, these findings validate our hypothesis that gradients from earlier iterations should be given less weight in the FS module.

\begin{figure*}[htbp]
\centering
\includegraphics[width=0.9\linewidth]{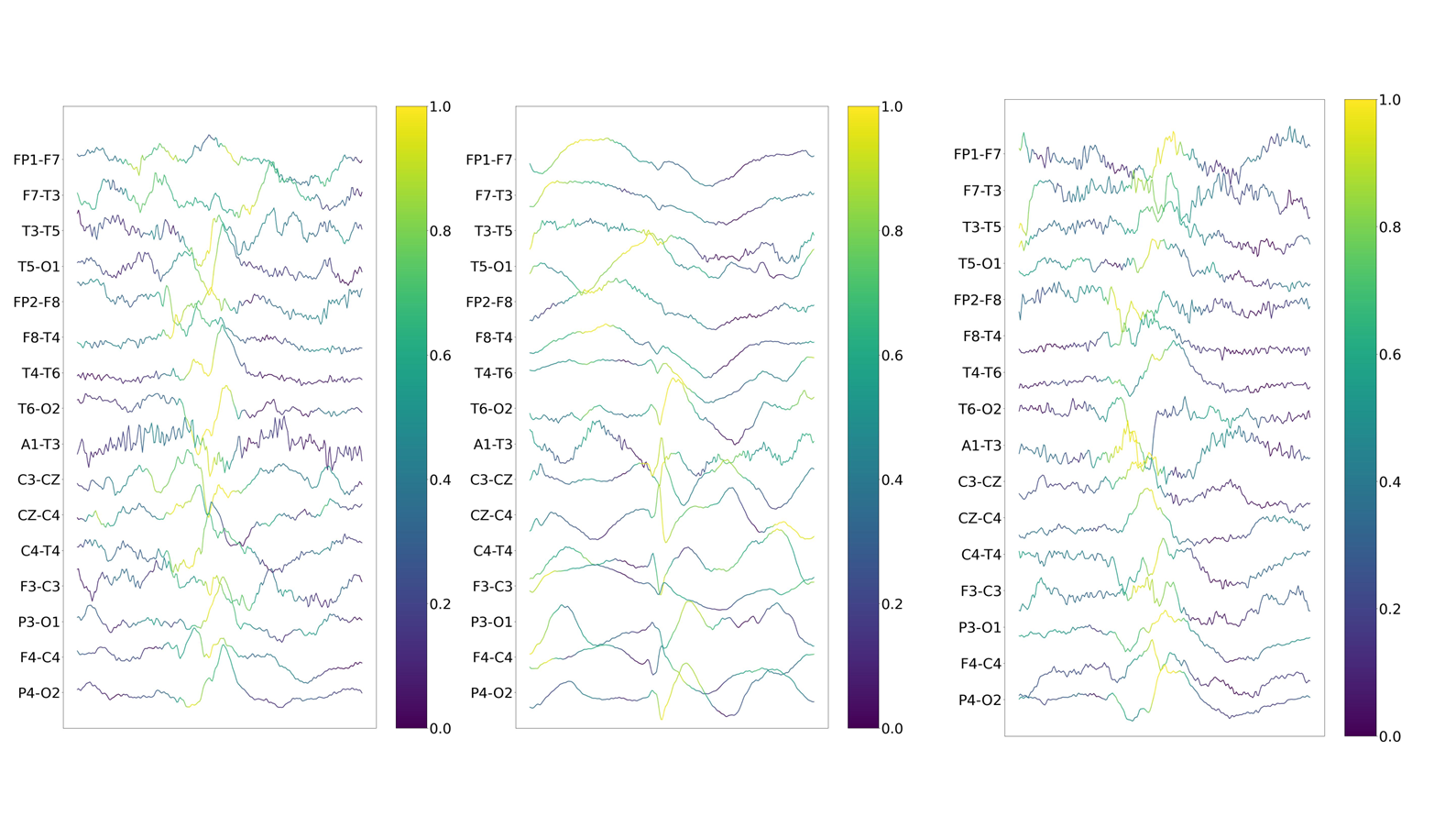}
    \caption{\textbf{Explainable analysis for EEG classification on TUEV.} We map the feature weights from our IEFS-GMB module to the corresponding raw EEG data for three samples in the TUEV dataset. The yellow areas in the three subfigures, where our model’s focus is concentrated, highlight the appropriate attention our model allocates.}
    \label{fig_explainable_analysis}
\end{figure*}
\noindent{\textbf{Explainable Analysis.}} In Figure \ref{fig_explainable_analysis}, we present a visualization of the feature weights produced by our IEFS module, derived for three distinct samples of the \textbf{TUEV} dataset. The interictal epileptiform discharges \cite{deJongh2005DifferencesIM}, which our model is designed to identify within the TUEV dataset, are marked by an an unusual clustering of peaks and troughs in adjacent channels, forming the notable “canine tooth” pattern. This pattern is clearly marked in yellow in the subfigures of Figure \ref{fig_explainable_analysis}, indicating where the model focuses most. Based on earlier studies \cite{ref_zhangliang,ref_hanyang,ref_kuntao}, we confirm that these visualizations demonstrate the ability of our model in pinpointing key EEG areas pivotal for IED detection.

\section{Conclusion}
This paper presents IEFS-GMB, a novel plug-and-play FS technique we’ve developed for EEG classification. It employs a gradient memory bank to preserve significant gradients drawn from earlier iterations, which are then used for subsequent information entropy calculations and FS. Experimental outcomes validate the efficacy of our method when combined with EEG encoders that exhibit the highest performance among eight SOTA models across four public EEG datasets. Moreover, IEFS-GMB surpasses four other SOTA FS methods. Extensive investigations into the robustness of IEFS-GMB and interpretability analyses highlight the method’s potential for practical clinical diagnostic applications.

\section*{CRediT authorship contribution statement}
\textbf{Liang Zhang:} Writing - original draft \& editing, Validation, Software, Methodology, Conceptualization, Formal analysis. \textbf{Hanyang Dong:} Writing - original draft \& editing, Validation, Software, Methodology, Conceptualization, Formal analysis.  
\textbf{Jia-Hong Gao:} Writing – review.
\textbf{Yi Sun:} Writing – review. \textbf{Kuntao Xiao:} Writing – review. \textbf{Wanli Yang:} Writing – review. \textbf{Zhao Lv:} Writing – review, Supervision. \textbf{Shurong Sheng:} Methodology, Writing – review \& editing, Formal analysis, Supervision. 

\section*{Declaration of Competing Interest}
The authors declare that they have no known competing financial interests or personal relationships that could have appeared to influence the work reported in this paper. The four EEG datasets used in our manuscript belong to public dataset. The patients involved in the database have obtained ethical approval. Users can download relevant data for free for research and publish relevant articles. Our study is based on open source data so there are no ethical issues and other conflicts of interest.

\section*{Acknowledgements}
This research was supported by the Distinguished Youth Foundation of Anhui Scientific Committee (No. 2208085J05), National Natural Science Foundation of China (NSFC) (No.62476004), Anhui Postdoctoral Scientific Research Program Foundation (No. 2024C894). 

\section*{Data availability}
The datasets employed in this study are publicly available. The code of our work will be released later. 

\bibliography{sample}

\end{document}